\documentclass[journal]{IEEEtran}

\usepackage{amssymb}

\usepackage{amsmath,amsfonts,bm}

\def\eqref#1{equation~\ref{#1}}

\def\1{\bm{1}}

\DeclareMathAlphabet{\mathsfit}{\encodingdefault}{\sfdefault}{m}{sl}
\SetMathAlphabet{\mathsfit}{bold}{\encodingdefault}{\sfdefault}{bx}{n}

\newcommand{\Cov}{\operatorname{Cov}}

\usepackage{cite}

\usepackage{hyperref}
\usepackage{url}
\usepackage{amsthm}
\usepackage{placeins}
\usepackage{verbatim}

\newcommand{\MMD}{\operatorname{MMD}}
\newcommand{\CORAL}{\operatorname{CORAL}}
\usepackage{algorithm}
\usepackage{algpseudocode}
\usepackage{graphicx}
\usepackage{caption}
\usepackage{subcaption}
\usepackage{booktabs}       

\usepackage{tikz}
\usetikzlibrary{arrows.meta,
                chains,
                positioning,
                shapes.geometric
                }

\tikzstyle{process} = [rectangle, minimum width=1.5cm, minimum height=1.2cm, text centered, draw=black, fill=blue!10]
\tikzstyle{arrow} = [thick,->,>=Stealth]

\title{Order Matters: Improving Domain Adaptation by Reordering Data}

\begin{document}

\author{Andrea Napoli, Paul White
\thanks{The authors are with the University of Southampton, UK (email: \{A.Napoli, P.R.White\}@soton.ac.uk).}
}

\maketitle

\begin{abstract}
Domain shift remains a key challenge in deploying machine learning models to the real world. Unsupervised domain adaptation (UDA) aims to address this by minimising domain discrepancy during training, but the discrepancy estimates suffer from high variance in stochastic settings, which can stifle the theoretical benefits of the method. This paper proposes Optimal Reordering of Data for Error-Reduced Estimation of Discrepancy (ORDERED), a novel unbiased stochastic variance reduction technique which reduces the discrepancy estimation error by optimising the order in which the training data are sampled. We consider two specific domain discrepancy losses (correlation alignment and the maximum mean discrepancy), formulate their stochastic estimation error as a function of the data sampling order, and propose a practical optimisation algorithm. Our simulations demonstrate reduced variance compared to related methods, and experiments on two domain shift image classification benchmarks show improved target domain accuracy.
\end{abstract}


\section{Introduction}

Machine learning models often underperform when the test data distribution differs from the training distribution, a phenomenon known as domain shift. Improving robustness to domain shift has been a longstanding goal in machine learning, and is crucial to the widespread deployment of AI \cite{Gulrajani2021InGeneralization,Koh2021WILDS:Shifts}.

Unsupervised domain adaptation (UDA) is a common approach, where models learn representations invariant across source and target domains by minimising a ``domain discrepancy" loss. Two widely used objectives are the correlation alignment (CORAL) loss, which matches covariance matrices \cite{Sun2016DeepAdaptation}, and the maximum mean discrepancy (MMD), which aligns kernel mean embeddings \cite{Tzeng2014DeepInvariance,pmlr-v37-long15,Li2018DomainLearning}. Although theoretically well-grounded \cite{Ben-David2006AnalysisAdaptation,Ben-David2010ADomains,Redko2022AGuarantees}, a key limitation is that empirically estimating these discrepancies is extremely noisy, especially in high dimensions and with small minibatches. This can destabilise training, and sometimes even lead to worse target domain performance than with no domain alignment at all \cite{Dubey2021AdaptiveGeneralization,Gao2023Out-of-DistributionAugmentations,Gulrajani2021InGeneralization,Koh2021WILDS:Shifts,Napoli2023UnsupervisedCalls,Napoli2024ImprovingSampling,Wang2019CharacterizingTransfer}.

The estimator noise can be lowered through the use of variance reduction, and this has previously been shown to improve performance in the UDA setting \cite{Napoli2024ImprovingSampling,Napoli2025VarianceSampling}. Although a large number of such techniques exist \cite{Defazio2014SAGA:Objectives,Johnson2013AcceleratingReduction,Shalev-Shwartz2013StochasticMinimization,Alain2015VarianceSampling,Johnson2018TrainingSampling,Katharopoulos2017BiasedTraining,Katharopoulos2018NotSampling,Kutsuna2025ExploringTraining,Loshchilov2016OnlineNetworks,Zhao2015StochasticMinimization,Peng2020AcceleratingSampling,Liu2018AcceleratingSampling,Zhang2017DeterminantalDiversification,Zhang2019ActiveProcesses,Bardenet2021DeterminantalSGD,Napoli2024ImprovingSampling,Napoli2024Diversity-BasedAdaptation,Zhao2014AcceleratingSampling,Liu2020AcceleratingStrata,Fu2017CPSG-MCMC:MCMC,Lu2021VarianceModels}, many require the loss to be additive over individual training examples, which renders them incompatible with UDA losses (which fundamentally depend on the interrelation between training examples). We defer to \cite{Napoli2025VarianceSampling,Gower2020Variance-ReducedLearning} for a full review of these techniques.

Our approach builds on \cite{Napoli2025VarianceSampling}, who reduce the variance via stratified
sampling \cite{Zhao2014AcceleratingSampling,Liu2020AcceleratingStrata}: the data are stratified using
discrepancy-specific clustering objectives, and minibatches are formed
by drawing a single instance uniformly and independently at random from
each stratum. Weighted loss functions are then used to correct for
imbalanced stratum sizes and ensure the losses remain unbiased.

This approach has three main shortcomings: 1) the strata are formed by
clustering based on a surrogate objective, which does not always
directly correspond to the estimator variance; 2) the strata are sampled
independently, which limits the degree of variance reduction which can
be achieved; 3) depending on the UDA loss, the corresponding clustering problem can be badly conditioned, resulting in low-quality solutions with small, singleton or even empty strata -- as well as having higher variance, this also slows down convergence of the training since it will take more steps to ``see'' all
the examples in the larger strata.

To address Shortcomings 1 and 2, our paper proposes an additional step
which directly and jointly optimises the sampling order of the data in each stratum.
This step minimises a new surrogate objective closer to the true
estimator variance. We call this method Optimal Reordering of Data for
Error-Reduced Estimation of Discrepancy (ORDERED). To address Shortcoming 3,
we also introduce a minimum cluster size constraint to precondition the clustering problem.

Deterministic sequencing of training data is a common area of research, though this is not always specifically to reduce variance. For example, curriculum learning \cite{Bengio2009CurriculumLearning,Wang2020ALearning} aims to present examples in increasing order of difficulty -- this reduces variance implicitly since harder examples have especially high gradient variance in the early stages of training. Other works directly optimise the data order or minibatch partitioning, for example using genetic algorithms \cite{Kumar2021ReorderingLearning}, entropy statistics \cite{Lu2022FantasticallySensitivity}, submodular optimisation \cite{Joseph2019SubmodularNetworks,Wang2019FixingPartitioning}, anti-clustering \cite{Papenberg2021UsingParts,Baumann2026AAlgorithm}, or the MMD \cite{Banerjee2021DeterministicNetworks}. The training distribution can also be varied using a weight schedule to mix multi-domain data \cite{Rukhovich2024CommuteLearning}.

Of these methods, ORDERED is most similar to anticlustering, which also optimises variance with respect to the minibatch permutation. However, whereas anticlustering only minimises the variance of the minibatch centroids, ORDERED explicitly minimises the variance of the UDA losses themselves.

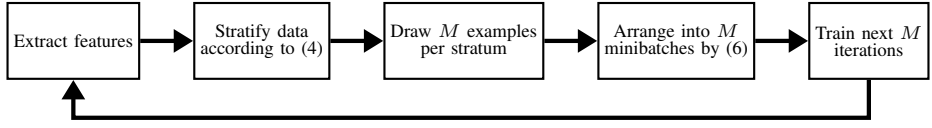
\begin{figure*}
\centering
\begin{tikzpicture}[
    node distance = 5mm and 7mm,
    start chain = going right,
    alg/.style = {draw, align=center, minimum height=1cm,  thick, font=\linespread{0.8}\selectfont},
    every join/.style = {-Triangle, ultra thick}
] 
\scriptsize

\begin{scope}[every node/.append style={on chain, join=by -Triangle}]
    \node (extract) [alg] {Extract features};
    \node (stratify) [alg] {Stratify data\\ according to (\ref{clusterobj})};
    \node (sample) [alg] {Draw $M$ examples\\ per stratum};
    \node (arrange) [alg] {Arrange into $M$\\ minibatches by (\ref{swapobj})};
    \node (train) [alg] {Train next $M$\\ iterations};
\end{scope}

\draw[-Triangle, ultra thick]
    (train.south) -- ++(0,-0.5)               
    -- ++(-10,0)                            
    -- ([yshift=-0.5cm]extract.south)        
    -- (extract.south);                    

\end{tikzpicture}
\caption{ORDERED training pipeline.}
\label{flowchart}
\end{figure*}

In the following sections, we introduce UDA variance reduction and the
stratified sampling approach of \cite{Napoli2025VarianceSampling}. We then formulate the stochastic estimation
errors of MMD and CORAL as a function of the data order, and
propose a practical optimisation algorithm. Using Monte Carlo simulations, we demonstrate significantly reduced variance for a given minibatch size, and show improved classification accuracy on two high-quality domain shift image
datasets.

\section{Method}\label{method}

\subsection{Preliminaries}\label{preliminaries}

Given labelled source examples \(x_{s,i},y_{s,i}\) indexed by
\(i \in \mathcal{I}_{s} = \left\{ 1,\ldots,n_{s} \right\}\), and
unlabelled target examples \(x_{t,j}\) indexed by
\(j \in \mathcal{I}_{t} = \left\{ 1,\ldots,n_{t} \right\}\), the goal of
UDA is to learn a model \(h\) that minimises some task loss
\(L_{\mathrm{task}}\) on the target domain. It is assumed \(h\)
decomposes into a featuriser \(f\) and prediction head \(g\), such that
\(h = g \circ f\). Since the target data are unlabelled, UDA methods
instead minimise \(L_{\mathrm{task}}\) on the source domain, alongside a domain
discrepancy loss \(L_{\mathrm{disc}}\) which aligns the source and target feature
distributions:
\begin{equation}\min_{h}{\mathbb{E}\left\lbrack L_{\mathrm{task}}\left( h\left( x_{s} \right),y_{s} \right) + \lambda L_{\mathrm{disc}}\left( f\left( x_{s} \right),f\left( x_{t} \right) \right) \right\rbrack},\end{equation}
where \(\lambda \in \mathbb{R}^{+}\) controls the trade-off between the
task and domain alignment objectives. This paper considers two specific
options for \(L_{\mathrm{disc}}\), the MMD and CORAL. The MMD is defined as
\begin{equation}L_{\MMD}\left( f\left( x_{s} \right),f\left( x_{t} \right) \right) = \left\| \mathbb{E}\left\lbrack \phi\left( f\left( x_{s} \right) \right) \right\rbrack\mathbb{- E}\left\lbrack \phi\left( f\left( x_{t} \right) \right) \right\rbrack \right\|_{\mathcal{H}}^{2}\end{equation}
where \(\mathcal{H}\) is a reproducing kernel Hilbert space, and
\(\phi :\mathcal{Z \rightarrow H}\) is an implicit mapping.
\(\mathcal{H}\) is associated with a unique positive-definite kernel
\(\kappa :\mathcal{Z \times Z}\mathbb{\rightarrow R}\) for which the
reproducing property
\(\kappa(z,z') = \left\langle \phi(z),\phi(z') \right\rangle_{\mathcal{H}}\)
is satisfied. On the other hand, CORAL aims to minimise the (squared)
Frobenius distance between the source and target feature covariance
matrices:
\begin{equation}L_{\CORAL}\left( f\left( x_{s} \right),f\left( x_{t} \right) \right) = \left\| \Cov\left\lbrack f\left( x_{s} \right) \right\rbrack - \Cov\left\lbrack f\left( x_{t} \right) \right\rbrack \right\|_{F}^{2}.\end{equation}
At training iteration \(m\), we select index subsets
\(B_{s}^{(m)} \subseteq \mathcal{I}_{s}\) and
\(B_{t}^{(m)} \subseteq \mathcal{I}_{t}\), each of cardinality \(k\), and
construct minibatches
\(\mathcal{B}_{s}^{(m)} = \left\{ \left( x_{s,i},y_{s,i} \right)\ |\ i \in B_{s}^{(m)} \right\}\)
and
\(\mathcal{B}_{t}^{(m)} = \left\{ x_{t,j}\ |\ j \in B_{t}^{(m)} \right\}\).
These are then used to compute stochastic losses
\({\widehat{L}}_{\mathrm{task}}^{(m)}\) and \({\widehat{L}}_{\mathrm{disc}}^{(m)}\), and
update \(h\). Our aim is to reduce the discrepancy estimation error
\(\sum_{m}^{}\left( {\widehat{L}}_{\mathrm{disc}}^{(m)} - L_{\mathrm{disc}}^{(m)} \right)^{2}\)
over the course of the training, by optimising how \(B_{s}^{(m)}\) and
\(B_{t}^{(m)}\) are chosen.

\subsection{Method overview}\label{method-overview}

Unfortunately, \({\widehat{L}}_{\mathrm{disc}}^{(m)}\ \)and\(\ L_{\mathrm{disc}}^{(m)}\)
depend on the features at iteration \(m\), making it hard to optimise
them directly. However, they can be well-approximated using features
from previous iterations, so long as the loss surface is locally smooth
and the learning rate is sufficiently small \cite{Liu2020AcceleratingStrata}. Intuitively, it
can be assumed that features that are close to each other at iteration
\(m\) will still tend to be close at iteration \(m + 1\). Therefore,
future minibatches are predetermined in sets of \(M\), based on features
\(z_{s,i} = f\left( x_{s,i} \right),\ z_{t,j} = f\left( x_{t,j} \right)\)
extracted at the current training iteration.

We build ORDERED on top of stratified sampling \cite{Napoli2025VarianceSampling}. That is, we
first partition \(\mathcal{I}_{s}\) and \(\mathcal{I}_{t}\) each into
\(k\) strata, \(S_{1},\ldots,S_{k}\) and \(T_{1},\ldots,T_{k}\)
respectively. We then sample \(M\)-tuples
\({\widetilde{S}}_{h},{\widetilde{T}}_{h}\) uniformly at random from
each stratum, which will form the next \(M\) source and target
minibatches. Specifically, the \(m^\text{th}\) minibatches are defined as
\(B_{s}^{(m)} = \bigcup_{h}^{}{\widetilde{S}}_{h}^{(m)}\),
\(B_{t}^{(m)} = \bigcup_{h}^{}{\widetilde{T}}_{h}^{(m)}\), comprising
the \(m^\text{th}\) element from each tuple, and the tuple orderings jointly
minimise a surrogate discrepancy estimation error based on
\(z_{s,i},\ z_{t,j}\). This approach ensures that the losses over the
whole training remain unbiased. The overall training pipeline is shown
in Figure \ref{flowchart}.

\subsection{Stratification}\label{stratification}

We construct the strata using dynamically-weighted kernel k-means
clustering \cite{Napoli2025VarianceSampling}, with a minimum cluster
size constraint to address Shortcoming 3. For the MMD, the clustering
objective for \(\mathcal{I}_{s}\) is
\begin{equation} \label{clusterobj}
\arg{\min_{S_{1},\ldots,S_{k}}{\sum_{h = 1}^{k}{\left| S_{h} \right|\sum_{i \in S_{h}}^{}\left\| \phi\left( z_{s,i} \right) - \frac{1}{\left| S_{h} \right|}\sum_{i \in S_{h}}^{}{\phi\left( z_{s,i} \right)} \right\|_{\mathcal{H}}^{2}}}}\end{equation}
subject to \(\left| S_{h} \right| \geq n_{\min}\), and analogously for \(\mathcal{I}_{t}\). For CORAL, the
objective is of the same form, but uses the specific mapping
\(\phi_{c}(z) = (z - \overline{z}){(z - \overline{z})}^{T}\). These
objectives are derived from the variance expressions of
\({\widehat{L}}_{\mathrm{disc}}^{(m)}\), and are shown to be good surrogates for
minimising the true variances when the data are sampled independently
for each stratum and iteration \cite{Napoli2025VarianceSampling}.

(\ref{clusterobj}) can be solved in a similar manner to \cite{Napoli2025VarianceSampling}, using a
Lloyd's-style alternating optimisation algorithm \cite{Lloyd1982LeastPCM}. Specifically,
the algorithm alternates between 2 steps:
\begin{enumerate}
\def\labelenumi{\arabic{enumi}.}
\item
  \textbf{Distance Update:} Compute the distance matrix
  \(P \in \mathbb{R}^{n_{s} \times k}\) from each datapoint to the
  centroid of each cluster using the kernel trick.
\item
  \textbf{Dynamically Weighted Assignment:} Compute the one-hot cluster
  assignment matrix \(U \in \left\{ 0,1 \right\}^{n_{s} \times k}\) that
  assigns each point to exactly one of the \(k\) clusters.
\end{enumerate}

\(U\) is the solution to the quadratic program
\begin{gather}  \label{quadprog}
\begin{split}
    &\arg{\min_{U}{\sum_{i,h}^{}\left\lbrack U_{ih}P_{ih}\sum_{i}^{}U_{ih} \right\rbrack}} \\
    &\text{s.t.} \quad   0 \leq U_{ih} \leq 1, \
    \sum_{h}^{}U_{ih} = 1,\ \sum_{i}^{}U_{ih} \geq n_{\min}.
\end{split}
\end{gather}
Since the Hessian of (\ref{quadprog}) is indefinite in general, this problem is
nonconvex and thus finding the global minimum is NP-hard. Although the
problem as currently defined could be readily input to a gradient-based
interior point method (to find a local minimum), these have
\(O\left( \left( {kn}_{s} \right)^{3} \right)\) complexity, and are
impractical above a few hundred data points. Instead, we solve (\ref{quadprog}) using a
greedy heuristic in a similar manner to \cite{Napoli2025VarianceSampling}, but with an extra
condition to satisfy the cluster size constraints. The algorithm
constructs \(U\) incrementally row-by-row, weighting the clusters using
interim cluster size values. Indices are assigned freely while there are
sufficient remaining datapoints to satisfy the constraints, after which
point the possible allocations are restricted to clusters that do not
yet reach the minimum size. This algorithm runs in
\(O\left( kn_{s} \right)\) time, and is listed in Algorithm \ref{alg:assign}, where
\(R(x) = \left\{ \begin{array}{l}
x,\ x \geq 0; \\
0,\ x < 0
\end{array} \right. \) is the ramp function.

\begin{algorithm}
\caption{Constrained weighted cluster assignments}\label{alg:assign}
\begin{algorithmic}[1]
\Require \(P \in \mathbb{R}^{n_{s} \times k},\ n_{\min} \in \mathbb{N}^+ \)
\Ensure \(U \in \left\{ 0,1 \right\}^{n_{s} \times k},\ \sum_{h}^{}U_{ih} = 1\)
\State $U \gets 0_{n_s \times k}$
\State $n_{1},\ldots,n_{k} \gets 0 \qquad \triangleright$ Interim cluster sizes
\State $r \gets n_s \qquad \triangleright$ Number of remaining assignments
\ForAll{\(i \in \left\{ 1,\ldots,n_s \right\}\)}
    \State $H \gets \{ h \in \left\{ 1,\ldots,k \right\} :n_{h} < n_{\min}\ \mathrm{or}$$\ r \geq \sum_{h = 1}^{k}{R\left( n_{\min} - n_{h} \right)} \}$
    \State $h \gets \arg{\min_{h \in H}{P_{ih}\left( n_{h} + 1 \right)}}$
    \State $U_{ih} \gets 1$
    \State $n_h \gets n_h+1$
    \State $r \gets r-1$
\EndFor
\State \Return $U$

\end{algorithmic}
\end{algorithm}

\begin{figure} [t]
    \centering
    \includegraphics[width=0.6\linewidth]{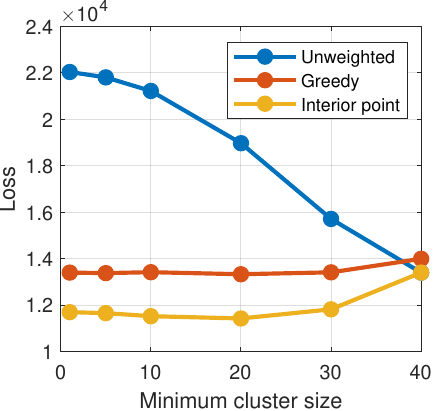}
    \caption{Objective value of (\ref{quadprog}) vs minimum cluster size $n_{\min}$ for three different optimisation algorithms.}
    \label{fig:nmin}
\end{figure}

Figure \ref{fig:nmin} shows how the loss attained by minimising (\ref{quadprog}) is affected by
the hyperparameter \(n_{\min}\). The input comprises Euclidean distances
between samples from a 2D standard normal distribution. We use a small
problem with \(n_{s} = 200\) and \(k = 5\), which allows us to compare
Algorithm \ref{alg:assign} with a commercial interior point solver \cite{TheMathWorksInc.2021MATLAB}. We also
test an unweighted constrained assignment, which is a linear problem and
can thus be solved quickly using linear programming, but does not
optimise the same objective. As expected, increasing \(n_{\min}\)
restricts the feasible problem space, which tends to increase the
achievable loss; however, the greedy algorithm appears less affected by
this than the interior point method. Note that as \(n_{\min}\)
approaches \(n_{s}/k\), the clusters tend to equal sizes, which is why
the unweighted optimiser approaches the weighted optimisers at this
point. It is also notable that the loss dips slightly for the interior point method around $n_{\min}=20$, which we posit is because the constraint is helping to precondition the assignment problem.

\subsection{Optimising sample order}\label{optimising-sample-order}

First, we present the sampling order optimisation problem in canonical
integer programming form. To proceed, let
\(\alpha \in \left\{ 0,1 \right\}^{n_{s} \times M},\beta \in \left\{ 0,1 \right\}^{n_{t} \times M}\)
be binary indicator variables for the source and target indices
respectively, such that \(\alpha_{im} = \left\{ \begin{array}{l}
1, \ i \in B_{s}^{(m)};  \\
0,\ \mathrm{otherwise}.
\end{array} \right.\), and likewise for \(\beta\). Define also cluster
size vectors
\(\mathbb{S} \in\mathbb{N}^{n_{s}},\mathbb{T} \in\mathbb{N}^{n_{t}}\),
where
\(\mathbb{S}_{i} = \left| S_{h} \right| \Leftrightarrow i \in S_{h}\)
(i.e., \(\mathbb{S}_{i}\) is the size of the cluster containing index
\(i\)), and equivalently for \(\mathbb{T}_{j}\), used to weight the
distance estimates to correct the sampling bias introduced by the
imbalanced clusters. Finally, let
\({\widetilde{B}}_{s} = \bigcup_{h}^{}{\widetilde{S}}_{h} = \bigcup_{m}^{}B_{s}^{(m)}\)
and
\({\widetilde{B}}_{t} = \bigcup_{h}^{}{\widetilde{T}}_{h} = \bigcup_{m}^{}B_{t}^{(m)}\)
be the union of all source and target indices for the next \(M\)
minibatches. The optimisation problem is thus
\begin{gather}
      \min_{\alpha,\beta}{\sum_{m = 1}^{M}\left( {\widehat{D}}^{(m)} - D_{0} \right)^{2}} \label{swapobj}\\
\hspace{-1.cm}\text{subject to} \ 
  \sum_{m}^{}\alpha_{im} = 1,i \in {\widetilde{B}}_{s},\sum_{m}^{}\beta_{jm} = 1,j \in {\widetilde{B}}_{t}\\
    \hspace{0.55cm} \sum_{m}^{}\alpha_{im} = 0,i \notin {\widetilde{B}}_{s},\sum_{m}^{}\beta_{jm} = 0,j \notin {\widetilde{B}}_{t}\\
    \sum_{i \in S_{h}}^{}\alpha_{im} = 1,\sum_{j \in T_{h}}^{}\beta_{jm} = 1,\\
    \alpha_{im},\beta_{jm} \in \left\{ 0,1 \right\},
\end{gather}
where \(D_{0}\) is the surrogate ``reference'' discrepancy over the full
dataset (approximating \(L_\mathrm{disc}^{(m)}\)), and \({\widehat{D}}^{(m)}\)
expresses the stochastic losses in terms of \(\alpha\) and \(\beta\)
(approximating \({\widehat{L}}_\mathrm{disc}^{(m)}\)). The reference (squared)
MMD is given by
\begin{equation}
D_{0,\MMD} = \left\| \frac{1}{n_{s}}\sum_{i = 1}^{n_{s}}{\phi\left( z_{s,i} \right)} - \frac{1}{n_{t}}\sum_{j = 1}^{n_{t}}{\phi\left( z_{t,j} \right)} \right\|_{\mathcal{H}}^{2},
\end{equation}
and the stochastic estimates are
\begin{equation}{\widehat{D}}_{\MMD}^{(m)} = \left\| \frac{1}{n_{s}}\sum_{i = 1}^{n_{s}}{\alpha_{im}\mathbb{S}_{i}\phi\left( z_{s,i} \right)} - \frac{1}{n_{t}}\sum_{j = 1}^{n_{t}}{\beta_{jm}\mathbb{T}_{j}\phi\left( z_{t,j} \right)} \right\|_{\mathcal{H}}^{2},
\end{equation}
or, in terms of kernel evaluations,
\begin{multline}
    {\widehat{D}}_{\MMD}^{(m)} = \frac{1}{n_{s}^{2}}\sum_{i,i' = 1}^{n_{s}}{\alpha_{im}\alpha_{i'm}\mathbb{S}_{i}\mathbb{S}_{i'}\kappa\left( z_{s,i},z_{s,i'} \right)} \\
 + \frac{1}{n_{t}^{2}}\sum_{j,j' = 1}^{n_{t}}{\beta_{jm}\beta_{j'm}\mathbb{T}_{j}\mathbb{T}_{j'}\kappa\left( z_{t,j},z_{t,j'} \right)} \\
 - \frac{2}{n_{s}n_{t}}\sum_{i,j = 1}^{n_{s},n_{t}}{\alpha_{im}\beta_{jm}\mathbb{S}_{i}\mathbb{T}_{j}\kappa\left( z_{s,i},z_{t,j} \right)}.
\end{multline}
The reference CORAL loss is
\begin{equation}D_{0,\CORAL} = \left\| C_{s,0} - C_{t,0} \right\|_{F}^{2},
\end{equation}
where \(C_{s,0}\) and \(C_{t,0}\) are the sample covariance matrices of
\(z_{s}\) and \(z_{t}\) respectively. The stochastic estimates are
\begin{equation}
    \begin{gathered}
{\widehat{D}}_{\CORAL}^{(m)} = \left\| {\widehat{C}}_{s}^{(m)} - {\widehat{C}}_{t}^{(m)} \right\|_{F}^{2} \\
{\widehat{C}}_{s}^{(m)} = \frac{1}{n_{s} - 1}\sum_{i = 1}^{n_{s}}{\alpha_{im}\mathbb{S}_{i}\left( z_{s,i} - {\widehat{\mu}}_{s}^{(m)} \right)\left( z_{s,i} - {\widehat{\mu}}_{s}^{(m)} \right)^{T}} \\
{\widehat{C}}_{t}^{(m)} = \frac{1}{n_{t} - 1}\sum_{j = 1}^{n_{t}}{\beta_{jm}\mathbb{T}_{j}\left( z_{t,j} - {\widehat{\mu}}_{t}^{(m)} \right)\left( z_{t,j} - {\widehat{\mu}}_{t}^{(m)} \right)^{T}} \\
{\widehat{\mu}}_{s}^{(m)} = \frac{1}{n_{s}}\sum_{i = 1}^{n_{s}}{\alpha_{im}\mathbb{S}_{i}z_{s,i}}, \quad
{\widehat{\mu}}_{t}^{(m)} = \frac{1}{n_{t}}\sum_{j = 1}^{n_{t}}{\beta_{jm}\mathbb{T}_{j}z_{t,j}} .
    \end{gathered}
\end{equation}
Although this problem could now be solved via standard integer programming methods, this will not be practical for large datasets. Instead, by considering the specific structure of the
problem, we propose a faster heuristic which searches for a local
minimum using a greedy strategy.

The approach begins with an initial random data order and reduces the
objective by iteratively swapping pairs of indices. Specifically, the
algorithm executes a single pass through the data, choosing the optimal
swap out of the \emph{remaining} elements in the same stratum via
exhaustive search. This means \(\frac{M(M - 1)}{2}\) objective
comparisons are performed per stratum, and thus \(kM(M - 1)\)
comparisons in total (for both \({\widetilde{B}}_{s}\) and
\({\widetilde{B}}_{t}\)). This algorithm is guaranteed to find a
permutation at least as good as the initial permutation. The algorithm
is listed fully in Algorithm \ref{alg:ordered}.

\begin{algorithm} []
\caption{ORDERED}\label{alg:ordered}
\begin{algorithmic}[1]
\State Initialise each $M$-tuple \({\widetilde{S}}_{1},{\widetilde{T}}_{1},...,{\widetilde{S}}_{k},{\widetilde{T}}_{k}\) with a random permutation

\ForAll{\(m \in \{1,\ldots,M\}\)}   $\qquad \triangleright$ Iteration index
    \ForAll{\(h \in \{1,\ldots,k\}\) } $\qquad \triangleright$ Stratum index

\State Swap elements \({\widetilde{S}}_{h}^{\left( m \right)}\) and
\({\widetilde{S}}_{h}^{\left( m_s \right)}\), where
\(m_s \in \left\{ m,\ldots,M \right\}\) and minimises (\ref{swapobj}).

\State Swap elements \({\widetilde{T}}_{h}^{\left( m \right)}\) and
\({\widetilde{T}}_{h}^{\left( m_{t} \right)}\), where
\(m_{t} \in \left\{ m,\ldots,M \right\}\) and minimises (\ref{swapobj}).

\EndFor
\EndFor
\State \Return \({\widetilde{S}}_{1},{\widetilde{T}}_{1},...,{\widetilde{S}}_{k},{\widetilde{T}}_{k}\)
\end{algorithmic}
\end{algorithm}

\begin{table*}[] \scriptsize
\centering
\caption{Average test accuracy for Spawrious by data split.}
\label{spawrious split}
\begin{tabular}{@{}l|cccccc|c@{}}
\toprule
\textbf{Method}      & \textbf{O2O-Easy}   & \textbf{O2O-Medium} & \textbf{O2O-Hard}   & \textbf{M2M-Easy}   & \textbf{M2M-Medium}  & \textbf{M2M-Hard}   & \textbf{Average}    \\ \midrule
ERM              & 68.6 ± 1.7          & 62.6 ± 0.8          & 62.1 ± 0.7          & 70.2 ± 1.8          & 45.0 ± 1.3           & 43.0 ± 1.2           & 58.6 ± 0.5          \\
DANN             & 91.4 ± 3.0          & 57.1 ± 3.5          & 71.1 ± 3.2          & 91.1 ± 0.1          & 54.8 ± 4.4           & 39.8 ± 3.4           & 67.5 ± 1.3          \\
CDAN             & 91.9 ± 1.7          & 57.0 ± 3.0          & 70.3 ± 2.2          & 92.9 ± 0.9          & 58.3 ± 3.8           & 44.3 ± 7.7           & 69.1 ± 1.6          \\
CDAN + SDAT      & 92.9 ± 1.4          & 54.3 ± 3.5          & 73.7 ± 6.6          & 83.3 ± 3.0          & 60.3 ± 4.3           & 53.0 ± 6.0           & 69.6 ± 1.8          \\
CDAN + ELS       & 89.8 ± 1.9          & 58.4 ± 2.1          & 67.3 ± 1.5          & 89.9 ± 2.2          & 62.3 ± 2.3           & 56.1 ± 10.2          & 70.6 ± 1.9          \\
ARM              & 70.2 ± 2.9          & 58.6 ± 2.1          & 60.6 ± 0.2          & 68.7 ± 1.6          & 42.2 ± 1.8           & 41.7 ± 1.0           & 57.0 ± 0.7          \\
MCC              & 87.6 ± 1.9          & 51.0 ± 0.8          & 54.0 ± 6.3          & 77.5 ± 2.4          & 46.4 ± 0.5           & 42.7 ± 1.2           & 59.9 ± 1.2          \\ \midrule
CORAL            & 70.7 ± 2.3          & 58.4 ± 1.9          & 64.1 ± 0.6          & 78.6 ± 1.5          & 54.1 ± 1.2           & 49.2 ± 0.7           & 62.5 ± 0.6          \\
+ k-means++      & 82.8 ± 3.5          & 58.2 ± 2.4          & 61.4 ± 4.1          & 75.5 ± 2.7          & 54.7 ± 2.7           & 48.6 ± 1.1           & 63.5 ± 1.2          \\
+ DPP            & 79.8 ± 3.4          & 59.8 ± 2.3          & 67.8 ± 2.0          & 79.6 ± 2.3          & 58.5 ± 1.4           & 49.4 ± 1.9           & 65.8 ± 0.9          \\
+ Anticlustering & 89.1 ± 4.7          & 57.6 ± 3.3          & 73.7 ± 6.1          & 87.5 ± 2.2          & 51.4 ± 3.2           & 47.5 ± 2.5           & 67.8 ± 1.6          \\
+ VaRDASS        & \textbf{90.1 ± 2.2} & \textbf{62.2 ± 1.0} & \textbf{79.6 ± 2.8} & 77.0 ± 2.7          & 51.7 ± 1.6           & 45.8 ± 0.4           & 67.7 ± 0.8          \\
+ ORDERED        & 88.2 ± 2.2          & 61.6 ± 1.6          & 78.1 ± 3.5          & \textbf{84.1 ± 5.0} & \textbf{60.5 ± 2.6}  & \textbf{50.5 ± 2.1}  & \textbf{70.5 ± 1.3} \\ \midrule
MMD              & 79.2 ± 3.3          & 61.9 ± 1.2          & 65.5 ± 3.4          & 76.2 ± 3.4          & 55.3 ± 3.4           & 48.1 ± 0.7           & 64.4 ± 1.1          \\
+ k-means++      & 83.7 ± 6.3          & 58.6 ± 2.4          & 68.4 ± 4.5          & 79.3 ± 2.7          & 60.0 ± 3.0           & 52.5 ± 4.5           & 67.1 ± 1.7          \\
+ DPP            & 83.6 ± 4.5          & \textbf{62.9 ± 0.9} & 63.5 ± 3.2          & 79.1 ± 3.1          & 57.4 ± 4.0           & 45.6 ± 1.7           & 65.4 ± 1.3          \\
+ Anticlustering & 74.5 ± 6.0          & 61.7 ± 1.0          & 85.9 ± 5.7          & 75.4 ± 4.1          & 63.7 ± 10.8          & 44.6 ± 3.0           & 67.6 ± 2.4          \\
+ VaRDASS        & \textbf{94.2 ± 1.8} & 61.5 ± 1.4          & 72.7 ± 4.5          & 76.9 ± 4.3          & \textbf{75.9 ± 10.6} & 48.1 ± 5.1           & 71.6 ± 2.2          \\
+ ORDERED        & 93.5 ± 1.3          & 56.4 ± 2.4          & \textbf{85.1 ± 1.9} & \textbf{88.6 ± 0.8} & 70.5 ± 7.8           & \textbf{62.1 ± 10.9} & \textbf{76.1 ± 2.3} \\ \bottomrule\end{tabular}
\end{table*}

\begin{table*}[]
\caption{Average test accuracy for Office-Home by data split.}
\label{office split}
\centering
\begin{tabular}{@{}l|cccccccccccc|c@{}}
\toprule
\textbf{Method}  & \textbf{C-A}  & \textbf{A-C}  & \textbf{P-A}  & \textbf{A-P}  & \textbf{R-A}  & \textbf{A-R}  & \textbf{P-C}  & \textbf{C-P}  & \textbf{R-C}  & \textbf{C-R}  & \textbf{R-P}  & \textbf{P-R}  & \textbf{Average}    \\ \midrule
ERM              & 32.9          & 33.9          & 30.7          & 45.7          & 49.1          & 57.5          & 34.2          & 46.2          & 36.5          & 51.8          & 63.2          & 59.6          & 45.1 ± 0.4          \\
DANN             & 31.1          & 34.9          & 28.9          & 39.9          & 46.9          & 50.5          & 31.7          & 46.8          & 39.7          & 48.7          & 63.4          & 52.7          & 42.9 ± 0.4          \\
CDAN             & 36.8          & 31.2          & 30.0          & 40.5          & 47.3          & 52.9          & 33.3          & 44.4          & 42.6          & 49.3          & 60.7          & 54.8          & 43.7 ± 0.5          \\
CDAN + SDAT      & 36.0          & 36.1          & 30.9          & 41.9          & 48.0          & 54.5          & 37.8          & 45.7          & 43.5          & 50.5          & 66.9          & 55.7          & 45.6 ± 0.2          \\
CDAN + ELS       & 35.1          & 30.7          & 29.3          & 38.3          & 45.7          & 52.9          & 33.6          & 44.3          & 41.6          & 46.7          & 62.7          & 55.2          & 43.0 ± 0.3          \\
ARM              & 34.2          & 31.3          & 30.0          & 43.1          & 49.2          & 56.5          & 32.6          & 46.4          & 35.6          & 47.6          & 63.1          & 56.0          & 43.8 ± 0.2          \\
MCC              & 33.8          & 38.3          & 34.3          & 50.7          & 49.9          & 58.7          & 36.4          & 54.1          & 44.2          & 55.6          & 69.1          & 59.8          & 48.7 ± 0.3          \\ \midrule
CORAL            & 39.4          & 35.2          & 34.7          & 40.8          & 57.1          & 56.8          & 35.8          & 48.3          & 43.2          & 49.5          & 68.9          & 60.0          & 47.5 ± 0.1          \\
+ k-means++      & 41.5          & \textbf{40.3} & \textbf{40.0} & \textbf{46.1} & 53.3          & 56.7          & 40.1          & 51.0          & 45.6          & 55.7          & 70.6          & \textbf{64.1} & 50.4 ± 0.5          \\
+ DPP            & 40.8          & 36.0          & 36.9          & 40.5          & 54.5          & 56.3          & 35.8          & 48.7          & 43.5          & 49.6          & 68.9          & 61.0          & 47.7 ± 0.2          \\
+ Anticlustering & 39.2          & 37.2          & 36.8          & 40.4          & 56.5          & 56.8          & 38.4          & 49.9          & 45.4          & 54.2          & 70.1          & 62.6          & 49.0 ± 0.2          \\
+ VaRDASS        & 39.7          & 35.6          & 36.6          & 42.8          & 55.7          & 57.9          & 39.3          & 50.3          & 47.3          & 52.8          & \textbf{71.6} & 62.3          & 49.3 ± 0.4          \\
+ ORDERED        & \textbf{41.9} & 37.3          & 39.6          & 44.9          & \textbf{58.6} & \textbf{58.3} & \textbf{42.2} & \textbf{51.4} & \textbf{47.4} & \textbf{55.8} & 71.5          & 64.0          & \textbf{51.1 ± 0.4} \\ \midrule
MMD              & 32.4          & 35.4          & 31.5          & 47.0          & 49.5          & 55.7          & 32.2          & 48.4          & \textbf{41.7} & 49.3          & 66.6          & 56.0          & 45.5 ± 0.3          \\
+ k-means++      & 33.6          & 33.7          & 32.4          & 43.9          & \textbf{51.8} & 53.4          & 31.9          & 48.0          & 39.3          & \textbf{52.5} & 66.2          & 55.5          & 45.2 ± 0.2          \\
+ DPP            & 31.6          & 34.9          & 31.0          & 45.2          & 51.0          & 56.3          & \textbf{33.4} & \textbf{51.2} & 39.2          & 50.1          & \textbf{67.2} & 59.0          & 45.9 ± 0.4          \\
+ Anticlustering & 35.1          & 33.2          & 33.0          & 44.6          & 50.1          & \textbf{57.2} & 33.3          & 48.4          & 36.6          & 51.0          & 63.8          & 58.3          & 45.4 ± 0.5          \\
+ VaRDASS        & 33.8          & 33.5          & 32.4          & 45.8          & 50.4          & 57.0          & 32.5          & 49.8          & 37.7          & 51.8          & 65.3          & 58.8          & 45.7 ± 0.2          \\
+ ORDERED        & \textbf{35.9} & \textbf{36.4} & \textbf{33.2} & \textbf{47.1} & 48.7          & 55.7          & 32.7          & 49.8          & 39.0          & 52.3          & 66.2          & \textbf{59.8} & \textbf{46.4 ± 0.2} \\ \bottomrule
\end{tabular}
\end{table*}

\begin{figure} [t]
     \centering
     \begin{subfigure}[b]{0.23\textwidth}
         \centering
         \includegraphics[height=0.75\textwidth]{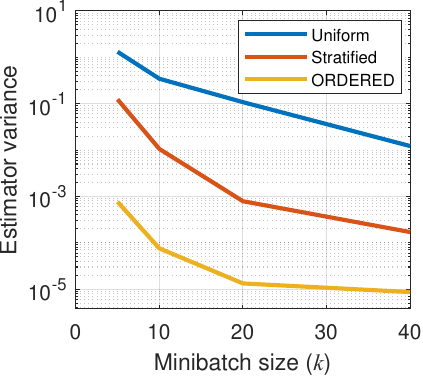}
         \caption{Estimator variance vs $k$ for ORDERED and two ablations.}
         \label{fig:3a}
     \end{subfigure}
     \hspace{.5cm}
     \begin{subfigure}[b]{0.16\textwidth}
         \centering
         \includegraphics[height=1.\textwidth]{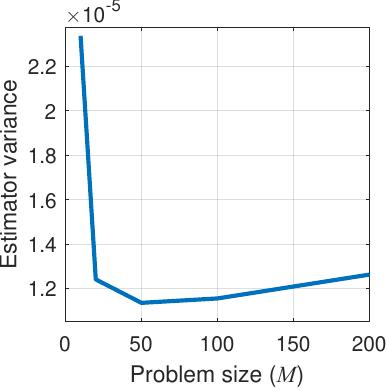}
         \caption{Estimator variance vs $M$ for ORDERED.}
         \label{fig:3b}
     \end{subfigure}
     \hfill
        \caption{The performance characteristics of Algorithm \ref{alg:ordered}.}
        \label{fig:3}
\end{figure}

We use Monte Carlo simulations to analyse the performance
characteristics of Algorithm \ref{alg:ordered} with respect to the parameters \(k\) and
\(M\). Specifically, we compute the variance of stochastic MMD estimates
using a linear kernel (that is, estimating the squared Euclidean
distance between distribution means) between a source and target dataset
comprising 2D standard normal data with \(n_{s} = n_{t} = 4,000\).
Figure \ref{fig:3a} compares the variance across different values of \(k\) for 3
samplers: uniform random sampling, stratified sampling \cite{Napoli2025VarianceSampling}, and ORDERED, with \(M = 100\). ORDERED achieves up to 2
orders of magnitude reduction in variance compared to stratified
sampling, and 4 orders of magnitude reduction compared to uniform random
sampling.

Figure \ref{fig:3b} shows how the variance changes with \(M\), with \(k = 20\). As
expected, the variance reduces significantly at first, since the
optimisation has greater degrees of freedom. However, perhaps
counter-intuitively, it can be seen to increase again for \(M > 50\). We posit
that this is because the smaller problem size induces more noise, which
helps to avoid local minima and achieve a better global solution.
Furthermore, for lower \(M\), the surrogate objective being optimised
\(\left( {\widehat{D}}^{(m)} - D_{0} \right)^{2}\) is closer on average
to the true deviation
\(\left( {\widehat{L}}_\mathrm{disc}^{(m)} - L_\mathrm{disc}^{(m)} \right)^{2}\),
which also improves reduction in variance. As well as the solution
quality, \(M\) is a trade-off in computational cost: lower \(M\)
requires more frequent extraction of features, but higher \(M\)
increases the complexity of Algorithm \ref{alg:ordered} quadratically.

Thus, the choice of \(M\) is influenced by a complex combination of
factors. For simplicity, we choose to fix \(M = 100\) for the remainder
of the experiments, which is the same update frequency chosen by
\cite{Napoli2025VarianceSampling}, and based on empirical observations from previous work
\cite{Liu2020AcceleratingStrata}.

\section{Experiments}\label{experiments}

In this section, the proposed method is evaluated in realistic training
conditions, to assess whether the observed reduction in variance
translates to an increase in test accuracy. Experiments are conducted
using the DomainBed framework \cite{Gulrajani2021InGeneralization} on the following domain shift
benchmarks.

\textbf{Spawrious} \cite{Lynch2023Spawrious:Biases} classification of 4 dog breeds
across images with different background environments (desert, jungle,
snow etc.). This benchmark comprises 6 data splits of varying difficulty, 6 domains and 18,664 examples.

\textbf{Office-Home} \cite{Venkateswara2017DeepAdaptation} image classification of 65 categories of everyday objects with different image styles (Art, Clipart, Product, and Real World). This benchmark comprises 12 data splits, 4 domains and 15,500 examples.

The domain discrepancies are measured between the union of all training
data and a held-out subset of the evaluation set. For the MMD, we use a
radial basis function (RBF) mixture kernel \cite{Li2018DomainLearning}, given by
\(\kappa(z,z') = \sum_{\gamma \in \mathcal{G}}^{}e^{- \gamma\left\| z - z' \right\|^{2}}\)with
\(\mathcal{G} = \{ 0.001,0.01,0.1,1,10\}\). For the clustering, we set
\(n_{\min} = M = 100,\) and sample
\({\widetilde{S}}_{h},{\widetilde{T}}_{h}\) without replacement, which
provides a further reduction in variance \cite{Gower2020Variance-ReducedLearning}.

The model comprises a pretrained ResNet-18 architecture \cite{He2015DeepRecognition}, which is
finetuned on the training data using the Adam optimiser \cite{Kingma2014Adam:Optimization} for 3,000
iterations. Hyperparameters, including the learning rate, weight decay, minibatch size $k$, and trade-off parameter $\lambda$, are tuned with a random search of size 10
using an in-distribution (training domain) validation set, independently
for each sampler. In particular, the random search distributions for $k$ and $\lambda$ are $k\sim2^{\text{Uniform}(3,7)}$ and $\lambda\sim10^{\text{Uniform}(-1,1)}$. The entire set of experiments is repeated 5 times for
reproducibility, using different random seeds for hyperparameters,
weight initialisations, and dataset splits. All other hyperparameter
choices and training details follow the DomainBed default options.

In total, five variance-reduced samplers are tested on the targeted UDA methods: k-means++ \cite{Arthur2007K-means++:Seeding,Napoli2024ImprovingSampling}, DPP \cite{Zhang2017DeterminantalDiversification,Napoli2024ImprovingSampling}, anticlustering \cite{Baumann2026AAlgorithm}, VaRDASS \cite{Napoli2025VarianceSampling}, and ORDERED. We also compare several baseline UDA methods: DANN \cite{Ganin2015Domain-AdversarialNetworks}, CDAN \cite{Long2017ConditionalAdaptation}, SDAT \cite{Rangwani2022ATraining}, ELS \cite{Zhang2023FreeSmoothing}, ARM \cite{Zhang2020AdaptiveShift}, and MCC \cite{Jin2020MinimumAdaptation}, plus non-adaptive training via ERM \cite{Vapnik1998StatisticalTheory}.

Tables
\ref{spawrious split} and \ref{office split} show the average test accuracy and
standard errors over the 5 repeats and each of the data splits, for
each method. The results confirm the importance of
effective variance reduction when estimating UDA losses.
ORDERED provides consistent increases in accuracy compared to other samplers, which allows the classical MMD and CORAL techniques to outperform significantly more modern UDA methods.

Overall wall-clock training times are reported for each sampler and dataset in Table \ref{times}. These are averaged over all data splits, hyperparameters, and repeats from the previous experiment, including both CORAL and MMD. It can be seen that the higher accuracy of ORDERED comes at a significant computational cost, being among the slowest of the methods compared and around an order of magnitude slower than simple random sampling.

\begin{table}[]
\caption{Average wall-clock training times by dataset (seconds).}
\label{times}
\centering
\begin{tabular}{@{}l|ccc@{}}
\toprule
\multicolumn{1}{c|}{\textbf{Sampler}} & \textbf{Spawrious} & \textbf{Office-Home} \\ \midrule
Uniform random                        & 393 ± 9            & 501 ± 18             \\
k-means++                             & 2312 ± 151         & 997 ± 28             \\
DPP                                   & 4123 ± 109         & 1095 ± 32            \\
Anticlustering                        & 1403 ± 24          & 818 ± 25             \\
VaRDASS                               & 2958 ± 21          & 1166 ± 45            \\
ORDERED                               & 3900 ± 83          & 2579 ± 97            \\ \bottomrule
\end{tabular}
\end{table}

\section{Conclusion}\label{conclusion}

This paper introduced ORDERED, a novel stochastic variance reduction
method for UDA based on reordering the training data. We showed that the
training data sampling order drastically influences the stochastic
estimation error of the MMD and CORAL losses, which in turn
significantly affects target domain performance. To address this, we
formulated the estimation error as a function of the data order, and
proposed a practical optimisation algorithm.

We believe the most promising direction for future work is in improving
the optimisation procedure, for instance by applying metaheuristics such
as simulated annealing or tabu search to enhance robustness against
local minima. The approach could also be extended to other UDA
objectives or a domain generalisation setting.

\section{Acknowledgements}\label{acknowledgements}

This work was supported by grants from BAE Systems and the Engineering
and Physical Sciences Research Council. The authors acknowledge the use
of the IRIDIS High Performance Computing Facility, and associated
support services at the University of Southampton, in the completion of
this work.

\bibliography{references.bib}
\bibliographystyle{ieeetr}

\end{document}